\documentclass[8pt,twocolumn, a4paper]{extarticle}

\usepackage{cite}
\usepackage[pdftex]{graphicx}
\usepackage{amsmath,amssymb}
\usepackage{wrapfig}

\usepackage[affil-it]{authblk}

\title{Point-wise Map Recovery and Refinement\\from Functional Correspondence}

\newcommand{\vct}[1]{\ensuremath{\mathbf{#1}}}
\newcommand{\T}{\ensuremath{^\top}}

\newcommand{\eg}{\emph{e.g.}}
\newcommand{\ie}{\emph{i.e.}}
\newcommand{\etal}{\emph{et al.}}

\author{Emanuele Rodol\`{a}\thanks{Email: \texttt{emanuele.rodola@in.tum.de}; Corresponding author}, Michael Moeller, and Daniel Cremers}
\affil{Department of Computer Science and Mathematics, Technische Universit\"at M\"unchen, Germany}
\date{}

\begin{document}


\maketitle

\begin{abstract}
Since their introduction in the shape analysis community, functional maps have met with considerable success due to their ability to compactly represent dense correspondences between deformable shapes, with applications ranging from shape matching and image segmentation, to exploration of large shape collections. Despite the numerous advantages of such representation, however, the problem of converting a given functional map back to a point-to-point map has received a surprisingly limited interest. In this paper we analyze the general problem of point-wise map recovery from arbitrary functional maps. In doing so, we rule out many of the assumptions required by the currently established approach -- most notably, the limiting requirement of the input shapes being nearly-isometric. We devise an efficient recovery process based on a simple probabilistic model. Experiments confirm that this approach achieves remarkable accuracy improvements in very challenging cases.
\end{abstract}

\section{Introduction}

Shape matching is a widely researched topic in computer vision and graphics, and a diverse range of techniques that tackle the problem of correspondence have been proposed during the years~\cite{vankaick11}. Of particular interest is the case in which the input shapes are allowed to undergo non-rigid deformations, which are typically assumed to be approximately isometric.  
Recent advancements in this area include the seminal work of Ovsjanikov~\etal \cite{ovsjanikov12}, who proposed modeling  {\em functional correspondence} between shapes; in this view, the focus shifts from studying point-wise mappings to the definition of a linear operator (the {\em functional map}) relating spaces of functions on the two shapes. The classical point-wise representation constitutes then a special case in which the functional map corresponds delta-functions to delta-functions. A major advantage of the functional representation lies in the linearity of the operator: the functional map admits a matrix representation which can be made {\em compact} under a  proper choice of bases for the two functional spaces. In \cite{ovsjanikov12} the authors advocated the use of the Laplacian eigenfunctions as the natural basis for smooth functions on the respective shapes; with this choice, one is allowed to ``truncate'' the representation by using the first few basis functions and still obtain a good approximation to the underlying point-wise correspondence.

\begin{figure}[h]
  \centering
  \includegraphics[width=0.9\linewidth]{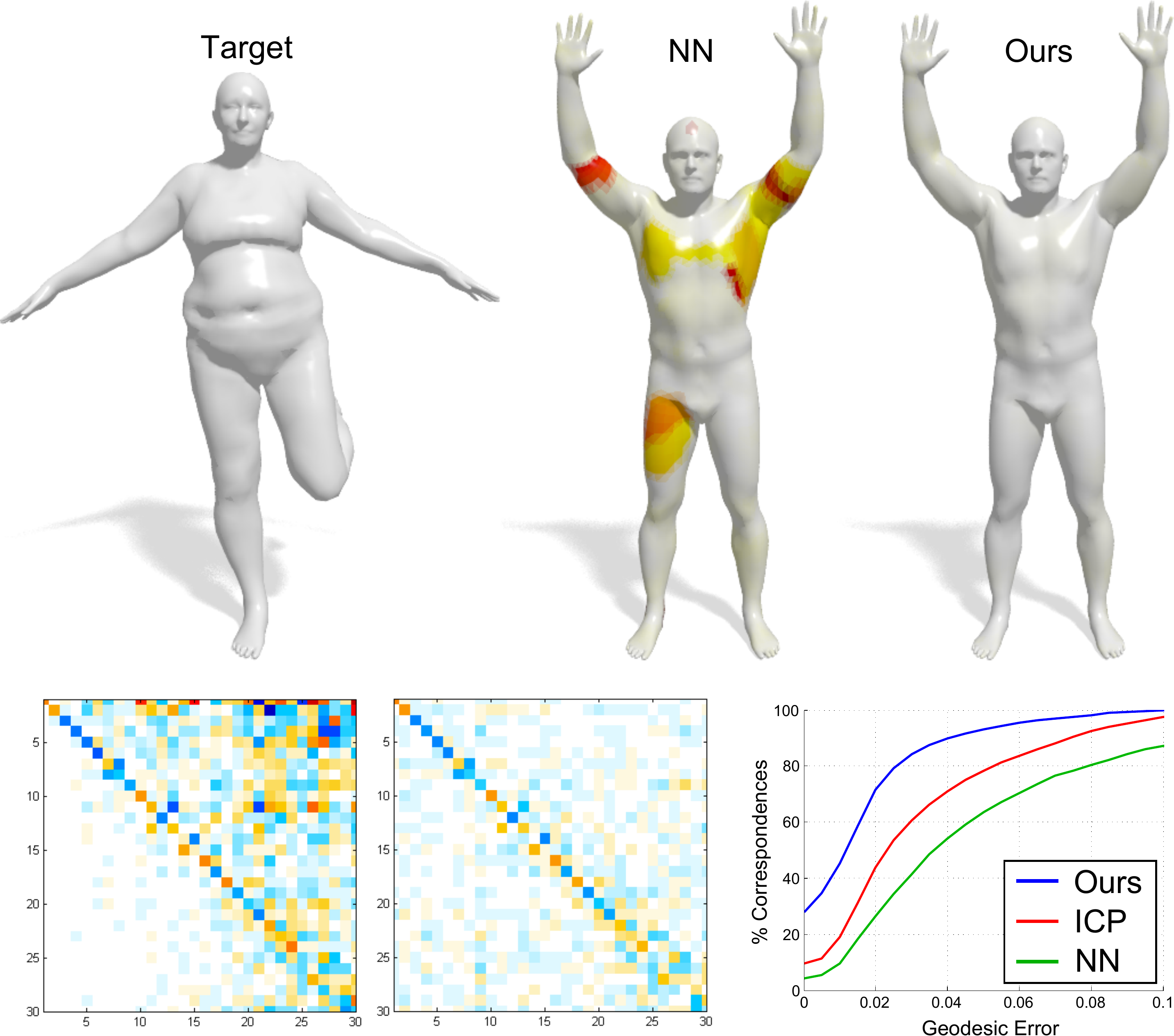}
  \caption{\label{fig:teaser}Given a functional map as input, our method allows to accurately recover and refine the underlying point-to-point mapping, even under non-isometric deformations. In the first row, color encodes distance to the ground-truth, increasing from white to red. The input map and its optimized version are shown in the second row.}
\end{figure}

Follow-up work by several authors showed how to extend the framework to deal with non-isometric deformations  \cite{pokrass13,kovnatsky13,rodola14,kovn15}, partial similarity \cite{rodola15}, shape exploration \cite{rustamov13,huan14} and image segmentation \cite{wang13} among others.
However, despite the success of these methods, there has been a general lack of interest on the inverse problem of accurately {\em reconstructing} a point-wise map from its functional representation -- a common requirement in many practical applications. 

The established approach, originally proposed in \cite{ovsjanikov12}, operates by formulating the conversion problem as a nearest-neighbor search in the embedded functional space; as we show in the following sections the approach works well if the functional map is accurate enough, with significant decrease in accuracy as the number of basis functions is reduced. The resulting point-to-point map can be iteratively refined by following a simple procedure, but this can only be done under specific assumptions on the initial functional map. A similar refinement technique was recently applied for near-isometric partial matching \cite{rodola15}, and in a correspondence-less setting for shape retrieval tasks \cite{gasparetto15}.
Finally, the point-wise recovery problem was sidestepped in \cite{kovn15} by adopting a soft error criterion to evaluate the quality of functional maps {\em without} converting them to a point-wise counterpart.

\paragraph*{Contribution.}
In this paper, we consider the problem of accurate point-wise map recovery from a given functional map. The key contributions can be summarized as follows:
\begin{itemize}
\item We provide the first rigorous analysis of the point-wise map recovery problem. In particular, we show how a simple modification to the baseline approach can lead to consistent improvements.
\item We introduce a simple probabilistic model for map recovery {\em and} refinement. Our model does not impose any assumption on the input functional map, as well as on the specific choice of a functional basis on the two shapes.
\item Our method is efficient, and significantly outperforms the existing method in both, the nearly-isometric and the inter-class settings.
\end{itemize}
%

\section{Background}
We model shapes as compact connected two-dimensional Riemannian manifolds $\mathcal{M}$ (possibly with boundary) endowed with the standard measure $\mu$ induced by the volume form. Let $L^2(\mathcal{M}) = \{ f: \mathcal{M} \rightarrow\mathbb{R} ~|~ \int_{\mathcal{M}}f^2d\mu <\infty \}$ denote the space of square-integrable functions on $\mathcal{M}$, and let $\langle f, g\rangle_{\mathcal{M}} = \int_{\mathcal{M}}fg d\mu$ be the standard manifold inner product.  The space ($\mathcal{M},\mu)$ features the symmetric Laplace-Beltrami operator (or Laplacian) $\Delta_{\mathcal{M}} : L^2(\mathcal{M}) \to L^2(\mathcal{M})$, which provides us with all the tools of Fourier analysis on our manifold. In particular, this  operator admits an eigen-decomposition ${\Delta_{\mathcal{M}} \phi_i = -\lambda_i \phi_i}$ for ${i\geq 1}$, with eigenvalues $0 = \lambda_1 < \lambda_2 \leq \hdots$ and eigenfunctions $\{\phi_i\}_{i\geq 1}$ forming an orthonormal basis on $L^2(\mathcal{M})$.

Drawing an analogy with classical signal processing theory, the eigenfunctions are often referred to as {\em manifold harmonics}, and the associated eigenvalues assume the interpretation of frequencies~\cite{taubin95}.
Any function $f\in L^2(\mathcal{M})$ then admits a Fourier series expansion as 
\begin{eqnarray}
\label{eq:fourier}
f(x) &=& \sum_{i\geq 1}  \langle f, \phi_i\rangle_{\mathcal{M}} \phi_i(x)\,.
\end{eqnarray}

\paragraph*{Functional correspondence.}
Let us be given two manifolds $\mathcal{M}$ and $\mathcal{N}$, and let $T:\mathcal{M}\to\mathcal{N}$ be a bijective mapping between them. Departing from the traditional point-centric setting, Ovsjanikov \etal \cite{ovsjanikov12} introduced the notion of functional map between two shapes as the linear operator $T_F: L^2(\mathcal{M}) \rightarrow L^2(\mathcal{N})$, mapping functions on $\mathcal{M}$ to functions on $\mathcal{N}$ via the composition
\begin{equation}
T_F(f) = f \circ T^{-1}\,.
\end{equation}
The approach is a natural generalization to classical point-wise correspondence, which can be seen as the special case in which $T_F$ maps indicator functions to indicator functions.

Let $\{\phi_i\}_{i\geq 1}$ and $\{\psi_i\}_{i\geq 1}$ denote orthonormal bases on $L^2(\mathcal{M})$ and $L^2(\mathcal{N})$ respectively. The functional correspondence with respect to these bases can be expressed as follows, for some function $f\in L^2(\mathcal{M})$:
\begin{eqnarray}
T_F(f) &=& 
T_F\left( \sum_{i\geq 1} \langle f, \phi_i \rangle_{\mathcal{M}} \phi_i \right)
= \sum_{i\geq 1} \langle f, \phi_i \rangle_{\mathcal{M}} T_F(\phi_i) \nonumber\\\label{eq:tf}
&=& \sum_{ij\geq 1} \langle f, \phi_i \rangle_{\mathcal{M}} 
\underbrace{\langle T_F(\phi_i), \psi_j \rangle_{\mathcal{N}}}_{c_{ij}} \psi_j\,.
\end{eqnarray}
Thus, the action of $T_F$ amounts to linearly transforming the expansion coefficients of $f$ from basis $\{\phi_i\}_{i\geq 1}$ onto basis $\{\psi_i\}_{i\geq 1}$.
The transformation is encoded in the coefficients $c_{ij}$, providing a representation for $T_F$ as the (possibly infinite) matrix $\mathbf{C} = (c_{ij})$. 
Seeking a functional correspondence among the two shapes then amounts to solving for the unknown $\vct{C}$ that better preserves certain mapping constraints~\cite{ovsjanikov12} or manifesting regularity at different levels~\cite{rodola14,kovn15}.

\paragraph*{Basis truncation.}

A natural choice for a basis on the two shapes is given by the eigenfunctions $\{\phi_i\}_{i\geq 1}$, $\{\psi_i\}_{i\geq 1}$ of the respective Laplacians (harmonic basis). The basis functions are said to be {\em compatible} among the two manifolds if the equality $\psi_i=\pm\phi_i \circ T^{-1}$ holds (approximately) for all $i\ge 1$, which is the case with the manifold harmonics when the shapes are related by a near-isometry.
If the deformation is far from isometric, compatible basis functions can still be computed ad-hoc for the two manifolds, based on a minimal set of coupling functions (\eg, a sparse set of point-to-point matches)~\cite{kovnatsky13}. In particular, since such basis functions still exhibit a natural ordering in the Fourier sense, they are said to form a {\em quasi}-harmonic basis.

Assuming to have stable and compact (namely, harmonic) bases at disposal, in \cite{ovsjanikov12} the authors proposed to truncate the series \eqref{eq:tf} at the first $k$ coefficients, resulting in a $k \times k$ matrix $\vct{C}$ approximating the full map in a compact way. 
The reduced representation greatly simplifies correspondence-based tasks (\eg, shape matching); at the same time, the truncation has the effect of `low-pass' filtering, thus producing smooth correspondences. In many applications, however, it is desirable to reconstruct the point-to-point mapping induced by the functional map. Thus, the interest shifts to the {\em inverse} problem of recovering the bijection $T$ from its functional representation $T_F$. 

\section{Point-wise map recovery}
\label{sec:recovery}

Let us now consider the discretized problem and assume that $\mathcal{M}$ and $\mathcal{N}$ are represented by a triangular mesh with $n$ nodes each. Let matrices $\Phi, \Psi \in \mathbb{R}^{n \times k}$ contain the first $k$  basis functions for the two shapes respectively, represented as column vectors. Note that, due to truncation, we have that $\Phi\T \Phi = \mathbf{I}_k$, but $\Phi \Phi\T \neq \mathbf{I}_n$, and similarly for $\Psi$. We assume that the bijection $T:\mathcal{M}\to\mathcal{N}$ is known. For the sake of simplicity, we additionally assume that $T$ can be represented as a permutation matrix $\vct{P} \in \{0,1\}^{n \times n}$, which means that the correspondence originates from two different deformations of the same template. In this case the expression for $c_{ij}$ in \eqref{eq:tf} can be equivalently rewritten as:
\begin{equation}\label{eq:ppp}
\vct{C} = \Psi\T \vct{P} \Phi\,,
\end{equation}
where $\vct{C}=(c_{ij})\in\mathbb{R}^{k \times k}$. Note that the matrix $\vct{C}$ is now a rank-$k$ approximation of $T_F$. The objective of any process converting the functional map back into a point-wise map representation, which is the focus of this work, is to recover the permutation $\vct{P}$ from the sole knowledge of $\vct{C}$, $\Phi$, and $\Psi$. 

\paragraph*{Assumptions.}
In order to be as general as possible, we do {\em not} assume the matching process which generated the given functional map to be known. Additionally, we do not require any particular properties from $\vct{C}$ (\eg, orthogonality), hence allowing to recover maps between shapes undergoing arbitrary deformations. Our only requirement is that the matrix representation $\vct{C}$ is given w.r.t. known bases $\Phi$, $\Psi$. These bases, in turn, may come from diverse optimization processes such as~\cite{pinkall93,kovnatsky13,neumann14}.

\paragraph*{Mapping indicator functions.}
\setlength{\columnsep}{7pt}
\setlength{\intextsep}{2pt}
\begin{wrapfigure}{r}{0.365\linewidth}
\begin{center}
\includegraphics[width=0.98\linewidth]{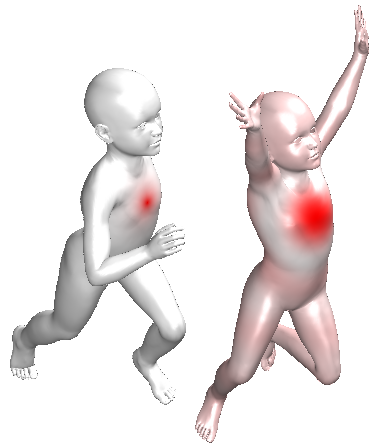}
\end{center}
\end{wrapfigure}
The simplest and most direct way for reconstructing the bijection $T$ from the associated functional map $T_F$ consists in mapping indicator functions $\delta_x:\mathcal{M}\to\{0,1\}$ for each point $x \in \mathcal{M}$ via $T_F$, obtaining the image $g=T_F(\delta_x)$, and then declaring $T(x)\in\mathcal{N}$ to be the point at which $g$ attains the maximum~\cite{ovsjanikov12}. Such a method, however, suffers from at least two drawbacks. First, it requires constructing and mapping indicator functions for all shape points, which can get easily expensive for large meshes. Second, the low-pass filtering due to the basis truncation has a delocalizing effect on the maximum of $g$ (see inset figure), negatively affecting the quality of the final correspondence.

\hyphenation{recovery}
\paragraph*{The inverse problem of point-to-point map recovery.}
Considering the problem of recovering $\vct{P}$ from a given $\vct{C}$ according to \eqref{eq:ppp} as a (highly underdetermined) ill-posed inverse problem, the natural recovery problem to consider is
\begin{align}
\label{eq:generalInverse}
\vct{P}^* = \arg \min_{\vct{P}\in \{0,1\}^{n\times n}}& D(\vct{C}, \Psi\T \vct{P} \Phi) + \alpha J^P(\vct{P})\\ \label{eq:doublyStochastic}
\mathrm{s.t.}~~&\vct{P}\T\vct{1}=\vct{1}\,,~\vct{P}\vct{1}=\vct{1}\,,
\end{align}
for a suitable measure of distance $D$, a regularization function $J^P$ to possibly impose some kind of desired smoothness of the transformation, and a regularization parameter $\alpha$ determining a trade-off between fidelity and regularity. 

Unfortunately, the minimization of \eqref{eq:generalInverse} will be extremely challenging in general. Consider for instance the case of no regularization, $J^P \equiv 0$, a zero functional map $\vct{C}=0$, and the distance measure being the squared Frobenius norm. In this case the minimization problem becomes
\begin{align}
 \min_{\substack{\vct{P}_{i,j}\in \{0,1\} \\ \vct{P}\T\vct{1}=\vct{1}\,,~\vct{P}\vct{1}=\vct{1}}} \|\Psi\T \vct{P} \Phi\|_F^2 &=
 \min_{\substack{\vct{P}_{i,j}\in \{0,1\} \\ \vct{P}\T\vct{1}=\vct{1}\,,~\vct{P}\vct{1}=\vct{1}}} \|\vct{Q} \vec{\vct{P}}\|_2^2 ,\nonumber \\
  &= \min_{\substack{\vct{P}_{i,j}\in \{0,1\} \\ \vct{P}\T\vct{1}=\vct{1}\,,~\vct{P}\vct{1}=\vct{1}}}  \vec{\vct{P}}\T \vct{Q}\T\vct{Q} \vec{\vct{P}} ,\nonumber
\end{align}
for the right hand sides denoting vectorized equations with $\vct{Q} = \Phi\T \otimes \Psi$ being formed by the Kronecker product between $\Phi\T$ and $\Psi$. For arbitrary $\vct{Q}$ the above problem is a particular reformulation of the quadratic integer programming problem (see equation (8) in \cite{bepp98}). Since the latter was shown to be NP-hard we cannot expect to find exact solutions to the above problem with reasonable computational complexity in general. Therefore, we turn our attention to approximations of \eqref{eq:generalInverse}. 

\paragraph*{Linear assignment problem.}
Instead of applying general greedy methods or relaxation techniques to \eqref{eq:generalInverse}, let us recall some observations from \cite{ovsjanikov12} regarding the general structure of \eqref{eq:ppp}:
In particular, note that the indicator function $\delta_x:\mathcal{M}\to\{0,1\}$ around point $x$ has coefficients $(\phi_i(x))_{i=1,\dots,k}$ in the Laplacian basis. This can be seen by observing that the inner product $\Phi\T\delta_x$ is  selecting the column of $\Phi\T$ corresponding to point $x$. Therefore, the image via $T_F$ of all indicator functions centered at points of $\mathcal{M}$ is simply given by $\vct{C}\Phi\T$. Recovering the point-to-point map could then be solved by finding the correspondence between the columns of ${\vct{C}\Phi\T}$ and the columns of $\Psi\T$. Measuring the proximity between these columns in the $\ell^2$ sense gives rise to the linear assignment problem
\begin{align}\label{eq:lap}
\min_{\vct{P} \in \{0,1\}^{n \times n}}~& \|\vct{C}\Phi\T - \Psi\T \vct{P}\|_F^2\\ \label{eq:doublyStochastic2}
\mathrm{s.t.}~~&\vct{P}\T\vct{1}=\vct{1}\,,~\vct{P}\vct{1}=\vct{1}\,.
\end{align}
Although the problem above admits a polynomial time solution~\cite{kuhn55}, typical values for $n$ (in the order of thousands) make computing this solution prohibitively expensive in practice.

\paragraph*{Nearest neighbors.}
The authors of \cite{ovsjanikov12} circumvent the computational costs of the above approach by proposing a nearest-neighbor technique for the recovery of the point-to-point correspondence. In the light of our previous analysis their idea is to consider the matching of every point, \ie, column of ${\vct{C}\Phi\T}$, to its nearest neighbor in $\Psi\T$ separately. 

Mathematically, the nearest-neighbor approach can be  seen as a relaxation of problem \eqref{eq:lap}, \eqref{eq:doublyStochastic2}, in which one seeks for the best {\em left-stochastic} approximation $\vct{P}$, \ie , 
\begin{align}\label{eq:nn1}
\min_{\vct{P} \in \{0,1\}^{n \times n}}~& \|\vct{C}\Phi\T - \Psi\T \vct{P}\|_F^2\\\label{eq:cl}
\mathrm{s.t.}~&\vct{P}\T\vct{1}=\vct{1}\,.
\end{align}
In other words, in comparison to \eqref{eq:doublyStochastic2} one omits the constraint of $\vct{P}\vct{1}=\vct{1}$. The omission allows to minimize the problem above by {\em independently} solving for columns of $\vct{P}$, one per query. The drawback of such a separable optimization approach, however, is that it may produce one-to-many mappings as a result of the recovery process. Moreover, the nearest-neighbor approach is an asymmetric method: looking for nearest neighbors from $\Psi\T$ to $\vct{C}\Phi\T$, or vice-versa, will in general produce different results.

\paragraph*{Balanced nearest neighbors.}
In order to remove the bias, we propose to incorporate additional terms in problem \eqref{eq:nn1}, namely minimize 
\begin{align}
&\|\vct{C}\Phi\T - \Psi\T \vct{P}\|_F^2 + \|\vct{C}\Phi\T \vct{Q} - \Psi\T\|_F^2+ \lambda \|\vct{P}-\vct{Q}\T\|_F^2 \nonumber \\ \label{eq:sym}
&\mathrm{s.t.}~\vct{P},\vct{Q} \in \{0,1\}^{n \times n}\,, \ ~\vct{P}\T\vct{1}=\vct{1}\,, \ ~\vct{Q}\T\vct{1}=\vct{1}\,,
\end{align}
where the minimization is performed w.r.t. both $\vct{P}$ and $\vct{Q}$.
Note that we incorporated the desired property of $\vct{P}$ being a permutation matrix by a soft constraint, \ie , by penalizing the difference of $\vct{P}$ to $\vct{Q}\T$. Also note that the limit of $\lambda \rightarrow \infty$ leads to a convergence of $\vct{P}$ to a solution of \eqref{eq:lap} meeting \eqref{eq:doublyStochastic2}. 

Instead of solving the minimization problem for increasing values of $\lambda$ exactly, we determine an approximate solution by alternating minimization on $\vct{P}$ and $\vct{Q}$. The latter leads to each subproblem being a simple nearest neighbor problem and guarantees the decrease of the objective functions. 

Table \ref{tab:alternating} illustrates the matching accuracy obtained by our symmetrized nearest-neighbor method in comparison to the classical nearest-neighbor as well as the indicator mapping approaches. As we can see, the symmetrization improves the results of the biased nearest-neighbor  method by 2--3$\%$. 
\begin{table}[t]
\centering
\begin{tabular}{cccc}
\hline
\# basis functions&\textbf{Max}&\textbf{NN}&\textbf{Balanced NN}\\
\hline
\textbf{25}&6.14&30.99&\textbf{33.24}\\
\textbf{50}&18.12&43.51&\textbf{45.65}\\
\textbf{75}&26.96&52.54&\textbf{55.07}\\
\hline
\end{tabular}
\caption{Percentage of exact matches recovered from functional maps of increasing rank. We show average results obtained by the three methods on 45 shape pairs from the FAUST dataset (intra-class, $\sim$7K vertices). The above comparison shows that the proposed balancing further improves the performance of the nearest neighbor technique.}
\label{tab:alternating}
\end{table}

\paragraph*{A probabilistic model.}
The analysis we provided above puts in evidence two major drawbacks, namely: 1) The linear assignment approach, the nearest-neighbor search, and our bi-directional variant, all rely on the assumption that the functional map $\vct{C}$ given as input aligns well the columns of $\Phi\T$ with those of $\Psi\T$ in the $\ell^2$ sense. 2) None of the above approaches incorporates regularity assumptions for the alignment process, \ie , the regularization term $J^P$ in the general inverse problem formulation \eqref{eq:generalInverse} was omitted. 

We propose to cast the point-to-point map recovery as a probability density estimation problem to obtain both, a better measure of proximity than the $\ell^2$ distance and a tool to impose regularity assumptions on the alignment map. 
Within our model, we interpret the columns of $\vct{C}\Phi\T$ as modes of a {\em continuous} probability distribution defined over $\mathbb{R}^k$ (the embedded functional space), while columns of $\Psi\T$ constitute the data, \ie, a discrete sample drawn from the distribution. The task is then to align the modes to the data, such that the point-to-point mapping can be recovered as the maximum posterior probability.

As a model for the distribution we consider a Gaussian mixture (GMM) with $n$ components, having the columns of $\vct{C}\Phi\T$ as centroids in $\mathbb{R}^k$. For simplicity, we assume the components have uniform weight $\frac{1}{n}$, and equal covariances $\sigma^2$. With this choice, the GMM density function is:
\begin{align}
p(\vct{y}) = \frac{1}{n} \sum_{i=1}^n p(\vct{y} | \vct{x}_i ; \sigma^2)\,,
\end{align}
where we write $\vct{y}$ and $\vct{x}_i$ for ${i=1,\dots,n}$ to denote the columns of $\Psi\T$ and $\vct{C}\Phi\T$ respectively, and define $p(\vct{y} | \vct{x}_i ; \sigma^2) = \frac{1}{(2\pi\sigma^2)^{k/2}} \mathrm{exp}( - \frac{\|\vct{y} - \vct{x}_i\|^2}{2 \sigma^2} )$.

Now let $R_\theta:\mathbb{R}^k \to \mathbb{R}^k$ denote the (unknown) transformation aligning the centroid locations $\vct{x}_i$ to the data points, according to a set of transformation parameters $\theta$. The alignment problem can then be solved by maximizing the likelihood, or equivalently by minimizing the negative log-likelihood function
\begin{align} \label{eq:like}
L(\theta,\sigma^2) &= -\sum_{j=1}^n \log\left( \frac{1}{n} \sum_{i=1}^n p(\vct{y}_j | R_\theta(\vct{x}_i) ; \sigma^2)\right)\,.
\end{align}
Note that the argument that minimizes \eqref{eq:like} can be also interpreted as the argument that minimizes the Kullback-Leibler (KL) divergence between a continuous GMM distribution (represented by $\vct{C}\Phi\T$) and a mixture of Dirac distributions (represented by $\Psi\T$). Hence, with our probabilistic model we are choosing the distance $D(\vct{C},\Psi\T\vct{P}\Phi)$ in Eq.~\eqref{eq:generalInverse} to be the (pseudo-)distance $D_\mathrm{KL}(\vct{C}\Phi\T,\Psi\T\vct{P})$.

Given optimal parameters $\theta$ and $\sigma^2$, the point-to-point correspondence probability between $\vct{x}_i$ and $\vct{y}_j$ can finally be obtained as the posterior probability $p(\vct{x}_i | \vct{y}_j)=\frac{1}{n} p(\vct{y}_j | R_\theta(\vct{x}_i);\sigma^2) / p(\vct{y}_j)$.

\paragraph*{Transformation function.}
The above  probabilistic model  leaves some freedom for the specific choice of a  transformation $R_\theta$. A simple example is choosing this transformation as a simple rotation, parametrized by $\theta$. To be more flexible we instead propose to consider the general transformation:
\begin{equation}\label{eq:displ}
R_\vct{V}(\vct{X}) = \vct{X} + \vct{V}\,,
\end{equation}
where parameters $\vct{V}$ assume the meaning of a displacement field, and $\vct{X} \equiv \vct{C}\Phi\T$. With this definition, the overall behavior of the transformation can be controlled by regularizing~$\vct{V}$.

\setlength{\columnsep}{7pt}
\setlength{\intextsep}{2pt}
\begin{wrapfigure}{r}{0.365\linewidth}
\begin{center}
\includegraphics[width=0.98\linewidth]{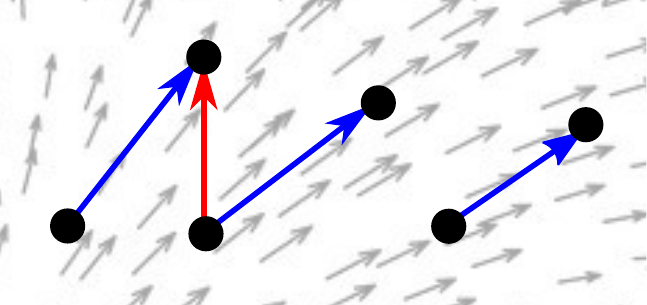}
\end{center}
\end{wrapfigure}

Assuming that the given functional map $\vct{C}$ represents a reasonably good matching, the refinement procedure should not be allowed to perturb the initial alignment significantly. Here we adopt the Tikhonov regularizer $\| \pmb \Gamma \vct{V} \|^2$ proposed in~\cite{yuille89,myronenko10}, where $ \pmb \Gamma$ is a low-pass operator promoting smoothly changing velocity vectors, hence coherent motion. In the inset we illustrate a smooth velocity field with coherent correspondences (blue) and a mismatch produced by simple nearest-neighbors (red).
We finally obtain the regularized objective:
\begin{equation}\label{eq:reg}
L(\vct{V},\sigma^2) + \lambda \| \pmb  \Gamma \vct{V} \|^2\,,
\end{equation}
where $\lambda > 0$ is a trade-off parameter between likelihood and regularity (set to $\lambda=3$ in our experiments).

General alignment problems like \eqref{eq:reg} have been widely researched in computer vision, and several robust algorithms exist for these tasks \cite{chui00,tsin04,myronenko10,jian11}. Most of these approaches follow an iterative scheme, optimizing w.r.t. $\{\theta,\sigma^2\}$ and $p(\vct{x}_i | \vct{y}_j)$ in an alternating fashion until convergence (EM algorithm \cite{dempster77}). In our experiments, we used publicly available code from~\cite{myronenko10}, which allows to optimize over smooth displacements as in Eq.~\eqref{eq:reg}.

\begin{figure}[t]
  \centering
  \includegraphics[width=0.95\linewidth]{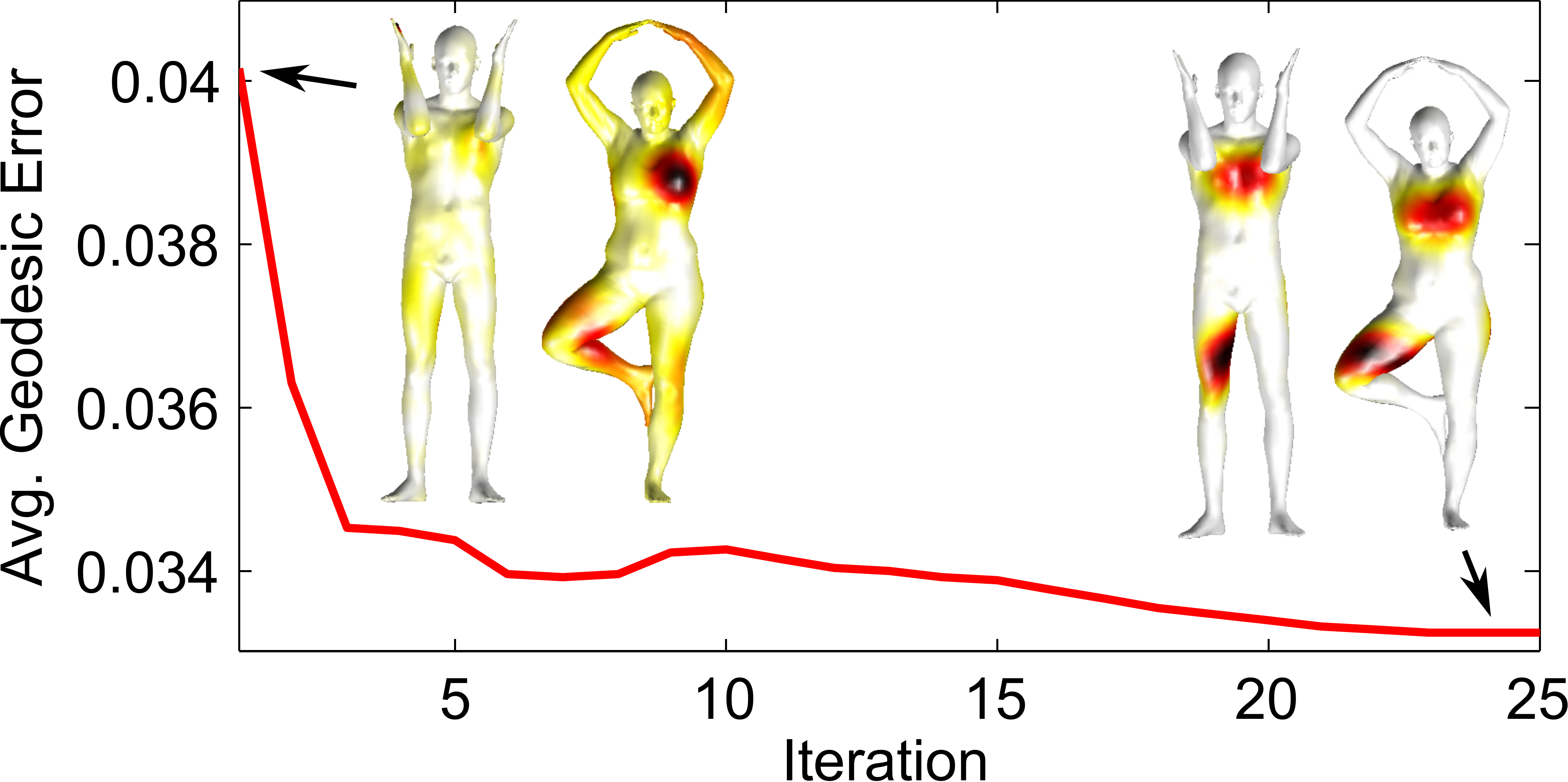}
  \caption{\label{fig:iterations}A few iterations of the minimization process for problem~\eqref{eq:generalInverse}. The curve shows the average geodesic error induced by the  point-wise correspondence obtained at each iteration. We visualize the quality of the initial and final maps by using the visualization technique of \cite{ovs13}. In this example, the ground-truth map exhibits a similar behavior to the one we show at the last iteration.}
\end{figure}

In Figure~\ref{fig:iterations} we illustrate a few iterations of the refinement process applied to a pair of nearly-isometric shapes,  
 starting from a functional map obtained by the matching algorithm described in Section~\ref{sec:experiments}. As a visual measure of map quality we employ the technique described in \cite{ovs13} (using the top 5 singular vectors), which allows to identify the problematic areas induced by a given functional map. The quality of a map can be judged by the smoothness of the associated plots, with better maps having a more localized behavior.

The output of the EM algorithm is a set of optimal transformation parameters and a left-stochastic correspondence $\vct{P}^\ast \in \{0,1\}^{n \times n}$.

%

\section{Point-wise map refinement}
\label{sec:refinement}
As a general representation for shape correspondences, functional maps can be adopted to compactly encode (via Eq.~\eqref{eq:ppp}) any dense point-to-point map obtained by generic matching algorithms. Therefore, one can consider locally refining the functional map $\vct{C}$ with the help of the point-to-point map recovered with any of the methods described in the previous section. Naturally, one can repeat such a strategy and iterate between updating the point-to-point correspondence and refining the functional map. 

\begin{figure}[t]
  \centering
  \includegraphics[width=0.95\linewidth]{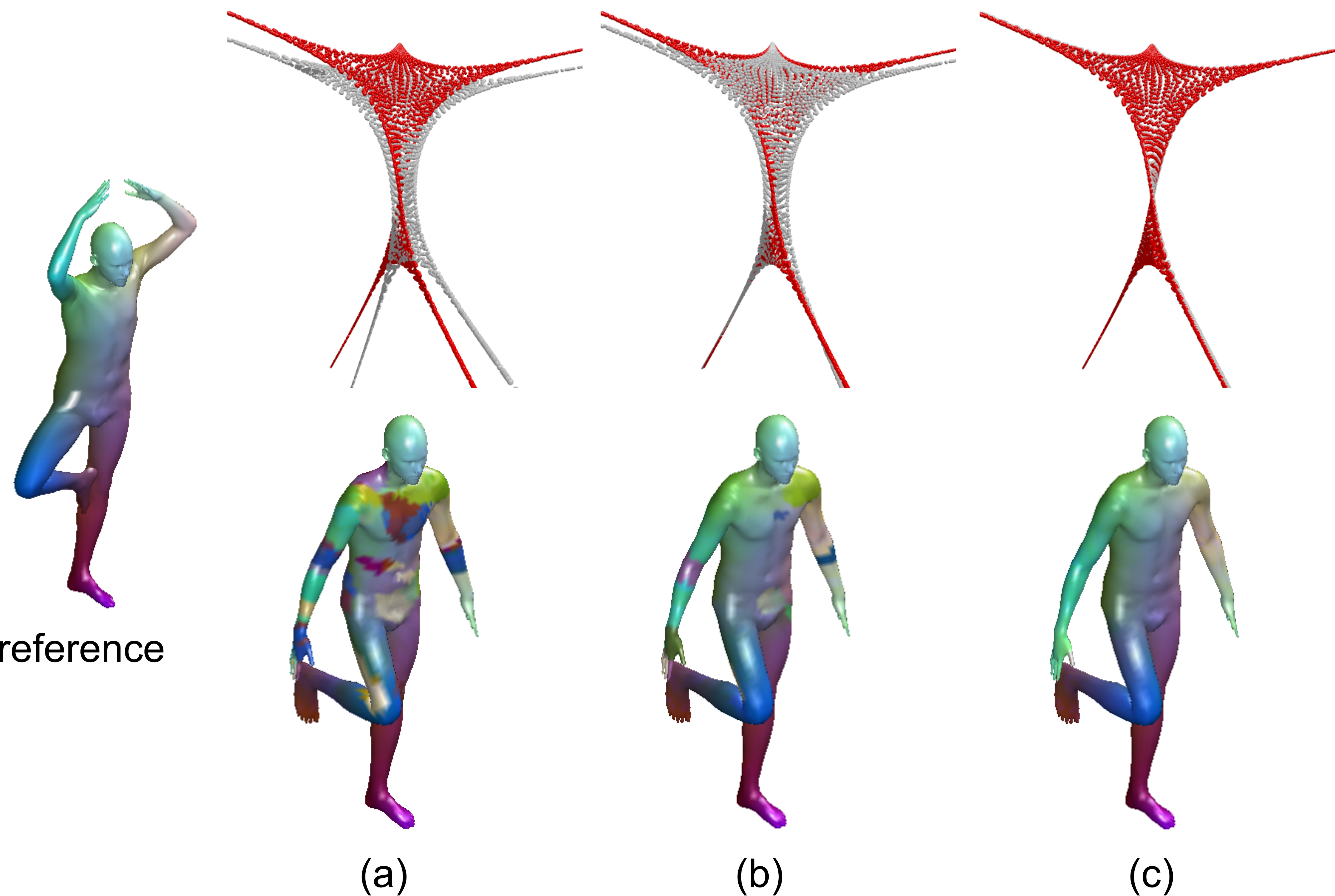}
  \caption{\label{fig:clouds}The refinement process of a rank-{\em k} functional map can be seen as the alignment of two point sets in $\mathbb{R}^k$. In the first row we illustrate the action of different refinement methods when $k=3$, with the two point sets being plotted as red and white point clouds. {\em (a)} Initial map; {\em (b)} Solution after orthogonal refinement \eqref{eq:icp}; {\em (c)} Solution obtained with our approach \eqref{eq:probp}. In the second row we show the color-coded point-wise matches obtained with each method when $k=50$.}
\end{figure}

Such an iterative procedure was considered in \cite{ovsjanikov12}, in which the authors proposed to use the classical Iterative Closest Point (ICP) algorithm~\cite{besl92}. The latter updates $\vct{P}$ according to the nearest-neighbor approach \eqref{eq:nn1}, followed by a refinement of $\vct{C}$ via
\begin{align}\label{eq:icp}
\min_{\vct{C}\in\mathbb{R}^{k\times k}}~& \|\vct{C}\Phi\T - \Psi\T \vct{P}\|_F^2\\\label{eq:ortho}
&\vct{C}\T\vct{C}=\vct{I}_k\,,
\end{align}
which is an orthogonal Procrustes problem. Intuitively, this can be seen as a rigid alignment in $\mathbb{R}^k$ between point sets $\Phi\T$ and $\Psi\T \vct{P}$ (see Figure~\ref{fig:clouds}(b) for an example). The alternating minimization w.r.t. $\vct{C}$ and $\vct{P}$ is repeated until convergence. 

Although the ICP approach described above allows to achieve significant improvements in terms of map accuracy, the orthogonality constraints \eqref{eq:ortho} imposed on the functional correspondence limit its applicability to a specific class of transformations, namely the class of volume-preserving isometries (see \cite{ovsjanikov12} Theorem 5.1). Therefore, the method cannot be applied to refine maps between shapes undergoing arbitrary deformations.

\begin{figure}[t]
  \centering
  \includegraphics[width=\linewidth]{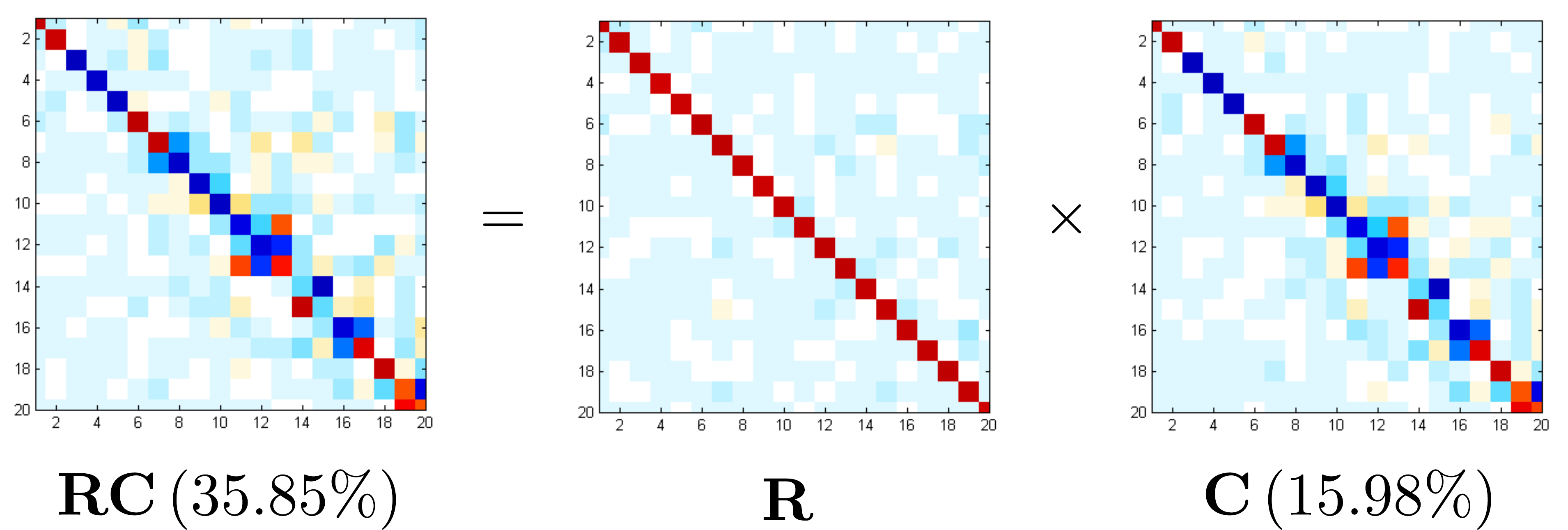}
  \caption{\label{fig:refinement}Solving problem \eqref{eq:probp} gives rise to the refined functional map shown on the left, where $k=20$. The refined map attains a higher percentage of exact matches (reported in parentheses) than the initial map. The optimal transformation (middle) essentially delineates the refinement process as a perturbation of the identity.}
\end{figure}

For a more general refinement procedure, we drop the orthogonality constraint on $\vct{C}$ and consider the problem
\begin{align}\label{eq:probp}
\min_{\vct{R}\in\mathbb{R}^{k\times k}}~& \|\vct{R}\vct{C}\Phi\T - \Psi\T \vct{P}\|_F^2  + \beta J^C(\vct{R}),
\end{align}
for a regularization functional $J^C$ which could encourage $\vct{R}\vct{C}$ to correspond to a smooth transformation, or could require $\vct{R}$ to be a small perturbation of the identity. While for the specific example 
$$ J^C(\vct{R}) = \left \{ \begin{array}{cc} 0 & \text{ if } (\vct{RC})\T\vct{RC} = \vct{I}_k, \\ \infty & \text{else,} \end{array} \right.$$
problem \eqref{eq:probp} coincides with the rigid alignment problem arising from the constraint \eqref{eq:ortho}, a less restrictive choice for the regularization functional $J^C$ makes the method suitable for recovering functional maps for non-isometric shape matching problems. 

In our experiments we found that when \eqref{eq:probp} is combined with our proposed regularized probabilistic model for recovering the point-to-point correspondence, it is sufficient to simply update $\vct{R}$ in a least-squares sense: Even without additional regularization, $\vct{R}$ is determined to be a perturbation of the identity as illustrated in Figure~\ref{fig:refinement}. The fact that $\vct{C}$ is refined in a {\em non-rigid} fashion can improve the refinement results significantly in comparison to orthogonal updates of $\vct{C}$ as illustrated in 3d in Figure~\ref{fig:clouds}.

%

\section{Experimental evaluation}
\label{sec:experiments}
We compare our iteratively refined probabilistic point-wise map recovery method with the iterative refinement procedure of Ovsjanikov~\etal \cite{ovsjanikov12} (denoted as {\textsc ICP}, see Section~\ref{sec:refinement}), which is to the best of our knowledge the only existing alternative to date. Both algorithms were implemented in Matlab/MEX and executed (single-core) on an Intel i7-3770 3.4GHz CPU with 32GB memory.

As a measure of error, we use the quantitative criterion that was introduced in~\cite{kim11} to evaluate the quality of point-wise maps. The input quantity in our case is a functional map $\vct{C}$, which is then converted to its point-wise counterpart using the two methods. We plot cumulative curves showing the percent of matches which have geodesic error smaller than a variable threshold. 

We evaluate the two methods quantitatively on the FAUST~\cite{bogo14} dataset, and qualitatively on the TOSCA~\cite{bbk08} and KIDS~\cite{rodola14} datasets. The three datasets include isometric as well as non-isometric shapes; in particular, FAUST and KIDS also include point-to-point ground truth matches between shapes belonging to different classes.

\paragraph*{Comparisons.}
The functional maps used in the comparisons are constructed by solving a least-squares system $\vct{C}\vct{A}=\vct{B}$, where matrices $\vct{A}$ and $\vct{B}$ contain the Fourier coefficients of indicator functions for corresponding regions on the two shapes. The region correspondence is established using the ground truth, while the sets of regions are computed using the consensus technique of \cite{rodola-cgf14}. This way, we simulate a matching process that provides reasonably good solutions for further refinement.

\begin{figure}[t]
  \centering
  \includegraphics[width=0.49\linewidth]{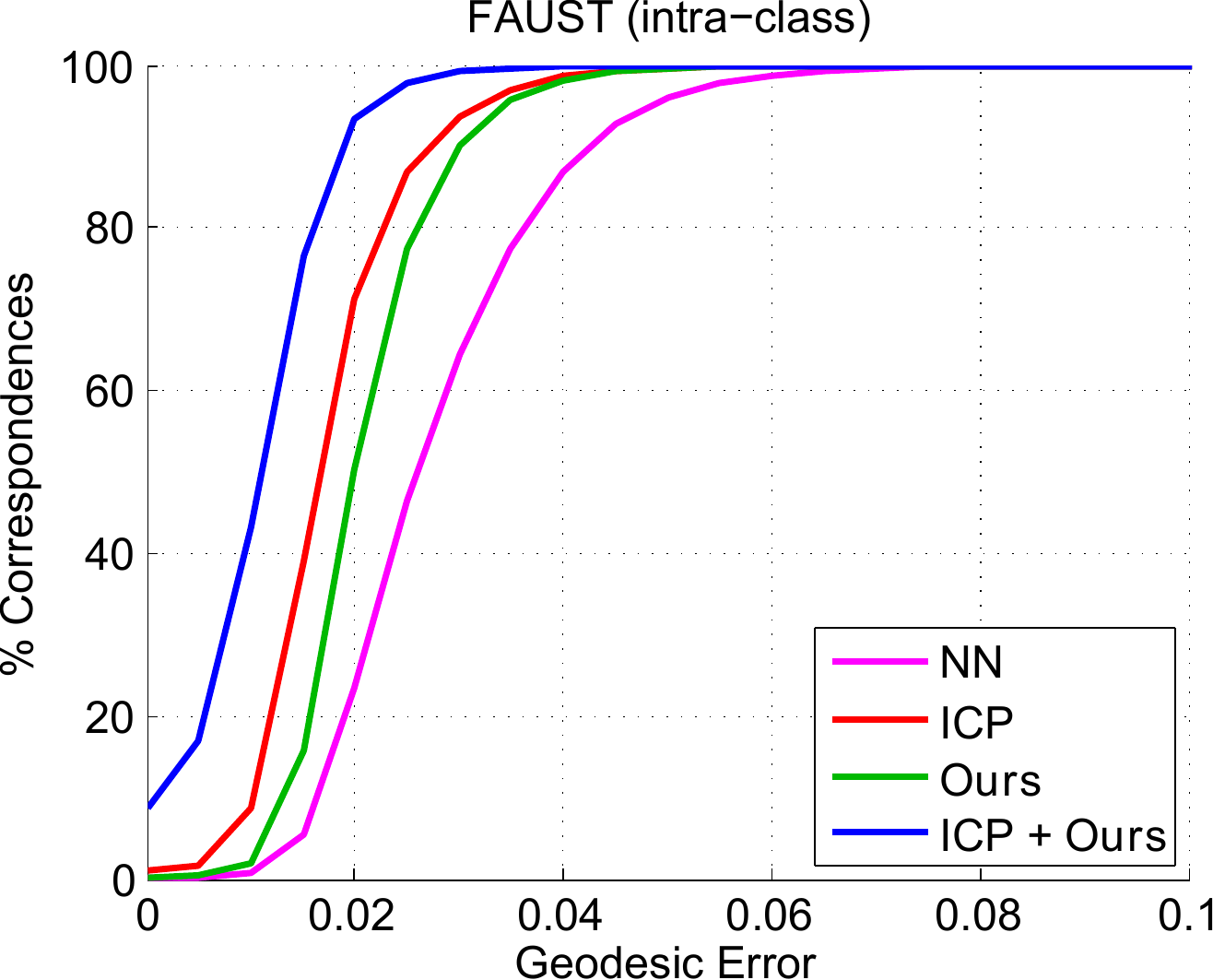}
  \includegraphics[width=0.49\linewidth]{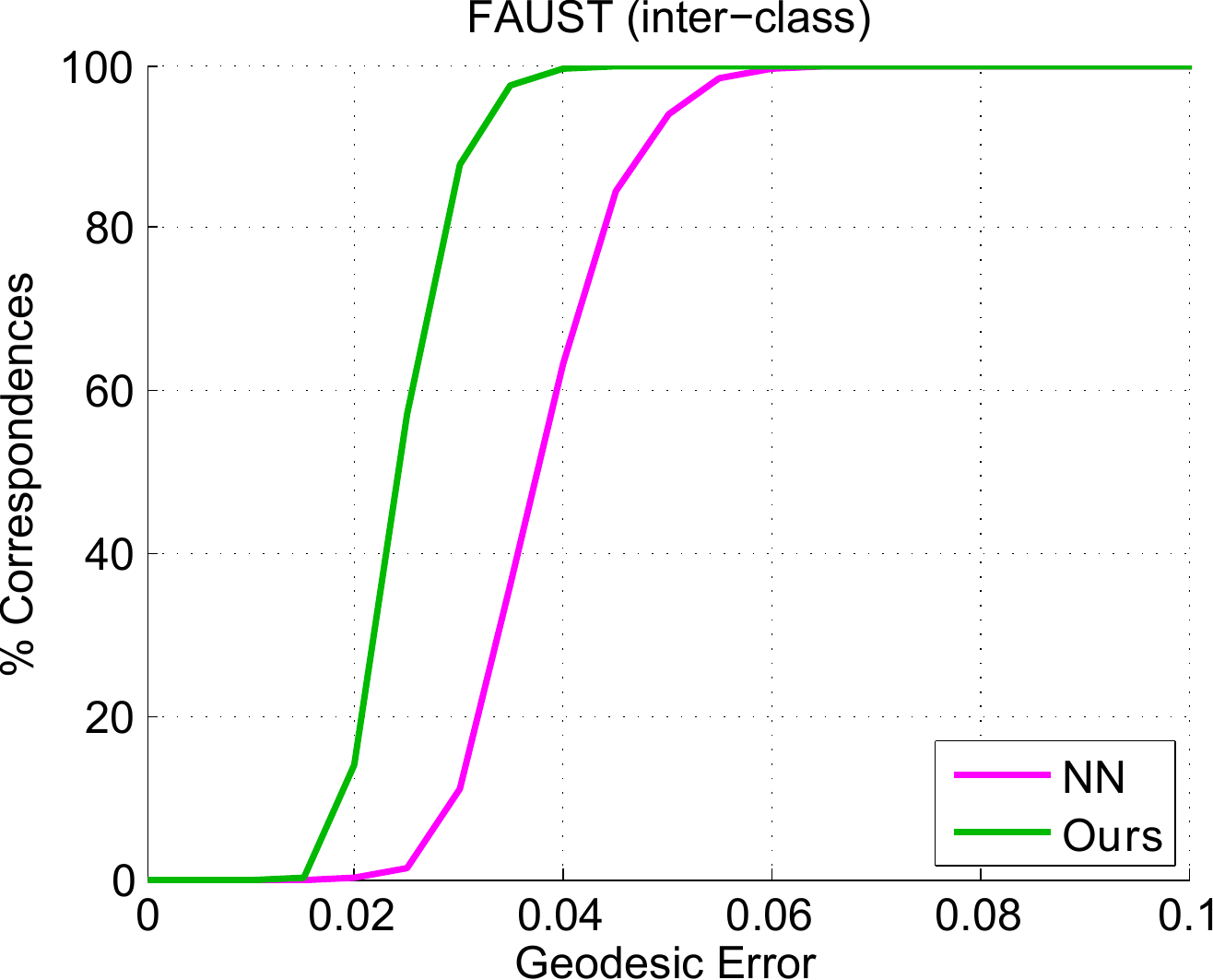}
  \caption{\label{fig:curves}{\em Left:} Comparisons with nearly-isometric shapes; in this case, orthogonal refinement already provides accurate point-to-point recovery, which can be further improved by applying our algorithm. {\em Right:} Comparisons with non-isometric shapes. In this case, ICP cannot be applied due to the lack of orthogonality in $\vct{C}$.}
\end{figure}

We show comparisons both in the near-isometric and inter-class settings. In the former case, we use 45 pairs of humans in different poses from the intra-class subset of FAUST. All shapes have $n=6890$ points, and the functional map is computed with $k=30$ basis functions. Results are reported in Figure~\ref{fig:curves} (left), where we also included plain nearest-neighbors ({\textsc NN}) as a baseline. From the plotted curves we can see that orthogonal ICP is performing slightly better than our method on near-isometric deformations, since the approach  is specifically tailored for this particular case. However, initializing our method with the output of ICP allows to achieve further $10\%$ improvement on average.

\begin{figure}[t]
  \centering
  \includegraphics[width=\linewidth]{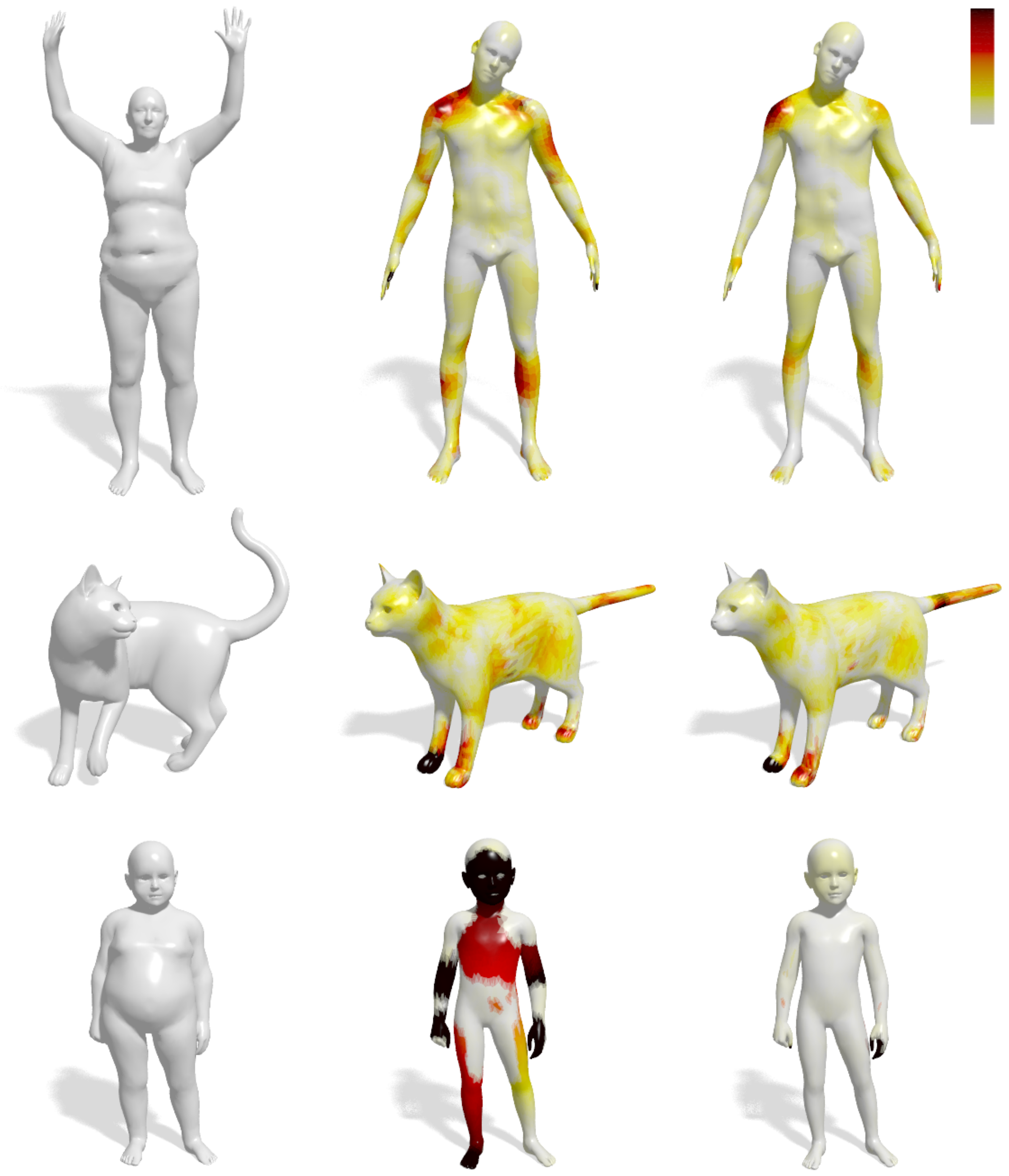}
  \caption{\label{fig:errormaps}Refinement examples in different matching scenarios. In each row we show the source shape (left), followed by the map errors produced by ICP (middle) and our method (right); the map error is visualized as a heat map encoding distance to the ground truth, growing from white to black. Note how the two methods perform comparably well in the near-isometric case (cat model), but orthogonal refinement yields large errors with more general deformations.}
\end{figure}

In Figure~\ref{fig:curves} (right) we show the same curves for the non-isometric case (inter-class matching). In this case, orthogonal refinement cannot be applied due to the different properties of the input functional maps, which relate shapes under {\em non} volume-preserving transformations.
Additional qualitative comparisons are shown in Figure~\ref{fig:errormaps}.

\paragraph*{Rank.}
In a second set of experiments, we evaluate the capability of the different methods to recover point-wise maps from functional maps of increasing rank. In this setting, we assume the input functional map to be as accurate as possible, and for this purpose we construct it explicitly as $\vct{C}=\Psi\T\vct{P}\Phi$, where $\vct{P}$ is the ground-truth permutation among the vertices of the two shapes. We do so for a pair of approximately isometric shapes, so that the respective eigenbases $\Phi$ and $\Psi$ are as repeatable as possible, and further orthogonal refinement is not needed (indeed, applying ICP in this setting actually yielded worse results in our tests). 

\begin{figure}[t]
  \centering
  \includegraphics[width=0.55\linewidth]{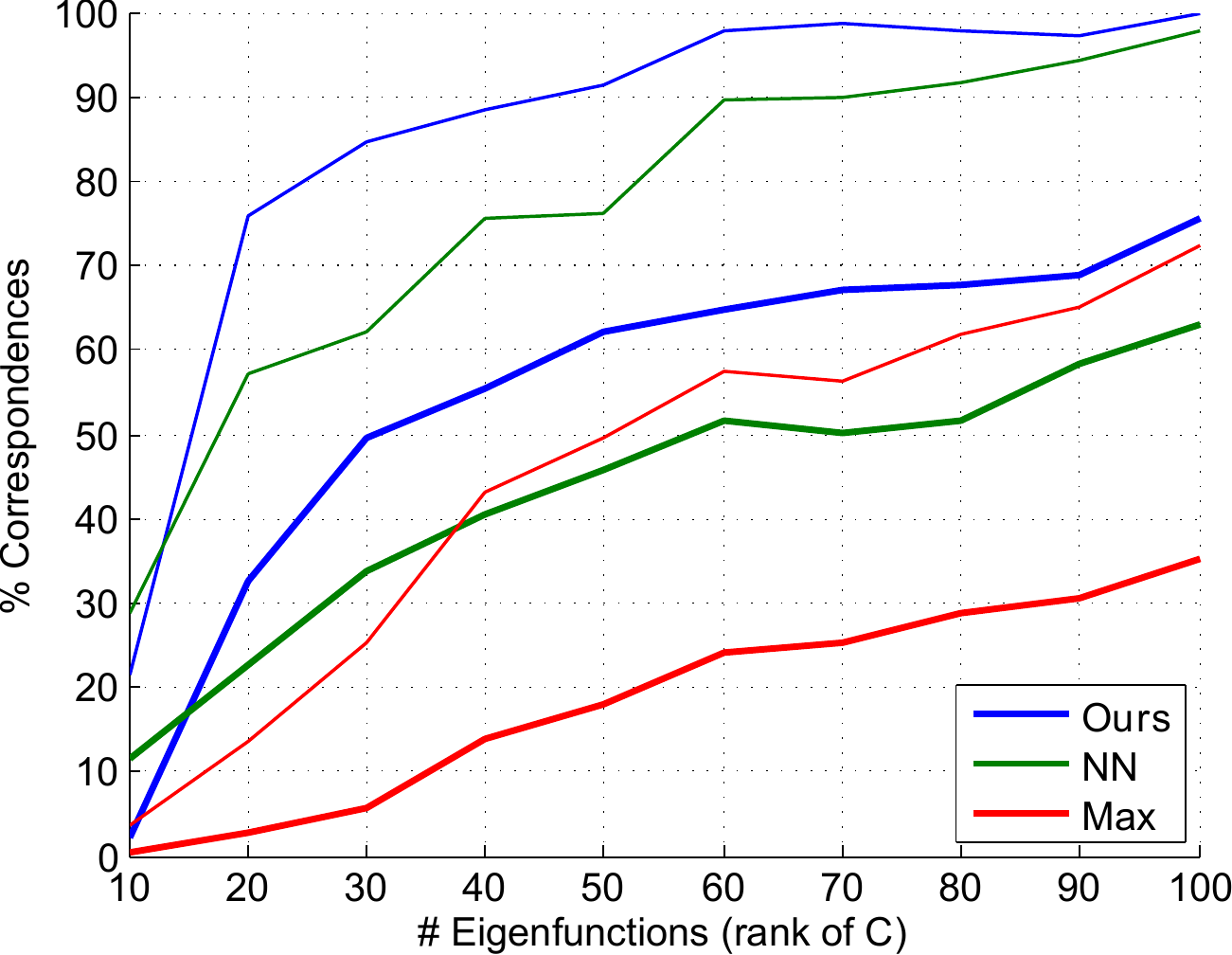}
  \includegraphics[width=0.43\linewidth]{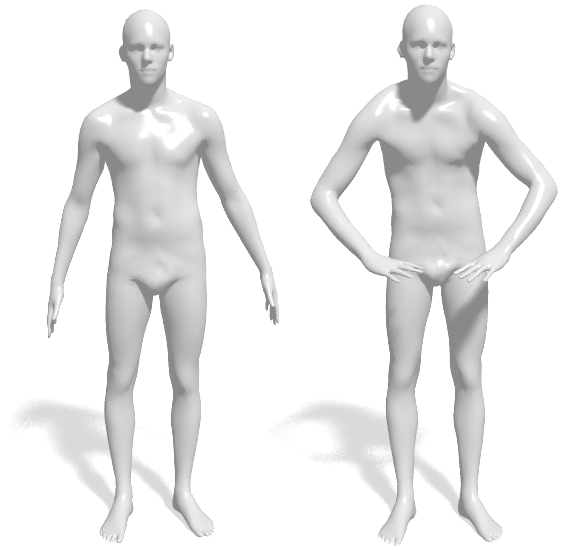}
  \caption{\label{fig:n_eigen}Percentage of exact correspondences (bold curves) recovered from ground-truth functional maps of increasing rank, among the two shapes shown on the right. We also report the percent of correspondences with geodesic error smaller than $0.02$ (thin curves).}
\end{figure}

The results are shown in Figure~\ref{fig:n_eigen}. As the number $k$ of basis functions used on the two shapes (\ie, rank of $\vct{C}$) increases, so does the amount of exact correspondences recovered by each method. This is also true for the simple indicator mapping approach ({\textsc Max}), since the smoothing effect due to basis truncation is reduced at increasing values of $k$. 

Our method allows to recover up to 20\% more {\em exact} matches than the nearest-neighbors approach. 
In particular, with ${k=100}$ (a commonly used value in most shape matching pipelines) we are able to perfectly reconstruct 75\% of the rows/columns of $\vct{P}$ (a $6890\times 6890$ matrix in this example).

\paragraph*{Complexity issues.}
The time performance of our method depends on two factors: the number of shape points $n$, and the size of the functional map $k$.
As we also show in Figure~\ref{fig:iterations}, typically a few iterations of the EM algorithm are sufficient to reach accurate solutions, and in practice we used 5 iterations in all our experiments.
In the typical case where $n=10,000$ and $k=50$, our method takes on average 1 min. to converge, while ICP using efficient search structures (kd-trees) adds up to $\sim$3 sec. 

It should be noted, however, that while we employed an off-the-shelf implementation of the minimization algorithm, this code can be easily parallelized and optimized in several ways. In particular, a GPU implementation of our method remains a practical possibility.

\section{Discussion and conclusions}
In this paper we considered the problem of point-to-point map recovery and refinement from its low-rank functional counterpart. Despite the growing success of this representation, the problem has received limited attention to date, and no significant steps have been made in this direction since the framework's conception. We formulated a general variational recovery approach for the inverse problem of computing point-to-point correspondences from a given functional map. We demonstrated what simplifications can be used to arrive at the nearest neighbor approach and showed how simply mitigating the asymmetry present in the standard conversion procedure can lead to consistent improvements of 2-3\% in accuracy. We then introduced a probabilistic model for point-wise map recovery and considered a refinement of the functional map that does not rely on the assumption of isometric shapes. The experimental results showed that the proposed approach yields up to 20\% accuracy improvements under non-isometric deformations, reaching up to 75\% {\em exact} point-to-point matches under good initializations.

The main limitation of our probabilistic method lies in the fact that -- similar to the nearest-neighbors approach -- the optimization procedure is biased towards one of the two shapes, as one can for instance see from the interpretation of minimizing the (non-symmetric) Kullback-Leibler divergence. Extending the ideas of the symmetrized nearest-neighbor approach to the probabilistic model for removing this bias represents a possibility. 
%
Second, while for simplicity we only considered pairs of shapes related by a bijection, our method can be modified to deal with shapes having different resolutions, as well as partially similar shapes and entire shape collections (joint refinement). We believe these topics to represent exciting directions of future research.

\section*{Acknowledgments}
We thankfully acknowledge Zorah L\"{a}hner and Federico Tombari for useful discussions. E.R. is supported by an Alexander von Humboldt fellowship.


\bibliographystyle{plain}

\bibliography{egbib}

\end{document}